\documentclass[a4paper,11pt,british]{article}
\usepackage{babel}

\usepackage[left=1.5in,right=1.5in,top=1.5in,bottom=1.5in]{geometry}

\usepackage{amsmath}
\usepackage{amsfonts}
\usepackage{xpatch}
\newcommand{\pvec}[1]{\vec{#1}\mkern2mu\vphantom{#1}}

\usepackage[ruled,vlined]{algorithm2e}
\SetKw{KwPuI}{Public Inputs:}
\SetKw{KwPrI}{Private Inputs:}
\usepackage{url}
\usepackage{mathtools}
\usepackage{booktabs}
\usepackage{siunitx}
\usepackage{physics}
\usepackage{bm}
\usepackage{orcidlink}
\usepackage{caption}
\usepackage[position=top]{subfig}
\usepackage{hyperref}

\makeatletter

\newcommand*\wthelper[2]{%
    \hbox{\dimen@\accentfontxheight#1%
        \accentfontxheight#11.2\dimen@
        $\m@th#1\widetilde{#2}$%
        \accentfontxheight#1\dimen@
    }%
}

\newcommand*\accentfontxheight[1]{%
    \fontdimen5\ifx#1\displaystyle
        \textfont
    \else\ifx#1\textstyle
        \textfont
    \else\ifx#1\scriptstyle
        \scriptfont
    \else
        \scriptscriptfont
    \fi\fi\fi3
}
\makeatother

\usepackage{amsthm}
\usepackage{thmtools}
\newtheorem{thm}{Theorem}[section]
\newtheorem{lem}[thm]{Lemma}
\newtheorem{cor}[thm]{Corollary}
\newtheorem{alg}[thm]{Algorithm}
\theoremstyle{definition}
\newtheorem{defn}[thm]{Definition}
\newtheorem{fact}[thm]{Fact}

\AtBeginDocument{\xpatchcmd\proof{$\blacksquare$}{\qedsymbol}{}{}}
\renewcommand{\qedsymbol}{}
\makeatletter
\renewenvironment{proof}[1][\proofname]{\par
  \pushQED{\qed}%
  \normalfont \topsep6\p@\@plus6\p@\relax
  \trivlist
  \itemindent\z@
  \item[\hskip\labelsep
        \scshape
    #1\@addpunct{.}]\ignorespaces
}{%
  \popQED\endtrivlist\@endpefalse
}
\makeatother

\begin{document}
\title{Towards Robust Federated Analytics via Differentially Private Measurements of Statistical Heterogeneity}

\author{Mary Scott\textsuperscript{\orcidlink{0000-0003-0799-5840}},
Graham Cormode\textsuperscript{\orcidlink{0000-0002-0698-0922}}, and
Carsten Maple\textsuperscript{\orcidlink{0000-0002-4715-212X}}}
\date{}

\maketitle

\begin{abstract}
Statistical heterogeneity is a measure of how skewed the samples of a dataset are.
It is a common problem in the study of differential privacy that the usage of a statistically heterogeneous dataset results in a significant loss of accuracy.
In federated scenarios, statistical heterogeneity is more likely to happen, and so the above problem is even more pressing.
We explore the three most promising ways to measure statistical heterogeneity and give formulae for their accuracy, while simultaneously incorporating differential privacy.
We find the optimum privacy parameters via an analytic mechanism, which incorporates root finding methods.
We validate the main theorems and related hypotheses experimentally, and test the robustness of the analytic mechanism to different heterogeneity levels.
The analytic mechanism in a distributed setting delivers superior accuracy to all combinations involving the classic mechanism and/or the centralized setting.
All measures of statistical heterogeneity do not lose significant accuracy when a heterogeneous sample is used.
\end{abstract}

\section{Introduction} \label{sec:intro}

Processing sensitive data in distributed environments, where the data is distributed across multiple devices, is becoming more common.
This is caused by the use of mobile devices increasing rapidly in the past decade or so, and companies such as Microsoft, Google, Apple and Dropbox advertising cloud storage options for ease of data access on all devices~\cite{wucloud}.
Too often, when analyzing data from multiple devices, an assumption is made that each local training dataset is independent and identically distributed (i.i.d.)~\cite{duan}.
However, in the real world most mobile devices differ greatly in terms of the apps stored, messages set, audio saved, etc.
Therefore, we need to find a way to measure and compensate for the difference, or statistical heterogeneity, in this data.

First, we give an overview of distributed environments, including the development of federated scenarios.
Federated learning (FL)~\cite{mcmahan} is an alternative to traditional centralized machine learning techniques where a single central server coordinates a large number of participating devices or \emph{clients} to provide results for its global model.
In each round, the server provides a subset of the clients with its current global model, after which each selected client uses its own local training dataset to generate results which it sends back to the server.
The server then aggregates these results, and uses the overall trends to improve its global model.
Federated analytics (FA)~\cite{ramage} complements FL by having the ability to measure, analyze and improve FL models, while simultaneously removing the need for the transfer or release of results from any individual client.
Real-world examples which use FA and FL include predicting the most likely word typed by a user given a sequence of the first few characters~\cite{zhu}, and the \emph{Now Playing} audio identification feature on Google's Pixel phones~\cite{yarcas}.

Although both of these federated scenarios attempt to protect the privacy of clients, via the use of decentralized training datasets and secure aggregation, there are no direct measures to perturb the raw data itself.
Therefore, for privacy purposes, FL and FA can be combined with differential privacy (DP), a mathematical definition that provides a strong, provable guarantee for the privacy of data.

Another problem with Federated Machine Learning systems such as FL and FA is that the accuracy of their models depends on the results being gathered from each small sample of clients being representative of those of the overall population.
However, in the real world this is not usually true: each individual typically has a particular style of messaging, a specific music taste and would use a select few favorite apps in their chosen way.
Therefore, the data of one client would give vastly different results for the word prediction and audio identification tasks described above, compared to the mean of a population.
Statistical heterogeneity (SH) is a term that can be used to describe a dataset if the observed trends from each sample differ from that of the overall population more than would be expected due to random error alone.

Therefore, it is likely that the general trends inferred from a statistically heterogeneous dataset will be less accurate than those from a non-SH dataset, unless there is a way to manipulate the way in which clients are chosen to favor those that will bring greater accuracy.
Furthermore, the combination of non-standard data with privacy is not as well studied as when an i.i.d. assumption can be put in place.
Our objective is to allow a quantitative measure of the degree of SH to be made.

In Section~\ref{sec:relwork}, we summarize the contributions that explore the ideas and themes of this work.
We look at the progress that has been made so far to tackle SH in the FA setting.
In Section~\ref{sec:prelims}, we give formal definitions of necessary concepts, such as DP and the measures of SH that we study.
In Section~\ref{sec:theory}, we build results for the mean squared error and a 95\% confidence interval from first principles for the measures of dispersion, $Q$ and $I^2$, using links between them to develop the more complex formulae in Sections~\ref{subsec:q} and \ref{subsec:isquared}.
In Section~\ref{sec:expts}, we present an experimental study to support these theoretical results.

\section{Related Work} \label{sec:relwork}

A definition of federated analytics (FA), adapted from Wang \emph{et al.}~\cite{wang}, is the collective execution of analytics tasks without raw data sharing, and the usage of the resulting information to create a desirable data environment via intelligent client selection.
This complements to the extensively studied federated learning (FL) framework, which relies on the complex training of a machine learning model, and the non-private sharing of weight updates with the servers.

As mentioned in Section~\ref{sec:intro}, the privacy of FA and FL can be improved significantly by incorporating differential privacy (DP)~\cite{dworkintro}, the requirement that the contribution of any single user to a dataset does not have a significant impact on the outcome of an analysis on that dataset.
DP can be achieved in a number of ways, the two most common being the centralized~\cite{dwork} and local~\cite{balleprivacyblanket} models of DP.
These contrasting models differ at the point where the \emph{random noise} is added: in the centralized model it is added by the central server after the data is collected, whereas in the local model each client perturbs their own data before sending it to the server.
The server not having to be trusted in the local model raises the level of noise required per user for the same privacy guarantee significantly, therefore limiting its practical usage to organizations with very large numbers of clients, such as Google~\cite{google}, Apple~\cite{apple} and Microsoft~\cite{microsoft}.

One of the main methods to guarantee DP in a centralized way in the non-federated setting is for the server to add noise from a Gaussian distribution.
The use of the Gaussian mechanism to guarantee DP can be extended to the FL framework, in particular to alter and approximate a federated optimization (FO) method, for example the averaging of weight matrices by a central server.
In FL, FO represents the clear optimization problem between both the intrinsically unbalanced nature of the network, as well as the numerous practical constraints such as client availability and behavior~\cite{mcmahan}.
Geyer \emph{et al.}~\cite{geyer} use the Gaussian mechanism to distort the sum of gradient updates so that a whole client's dataset can be protected, instead of a single data point.

One of the major drawbacks of using DP is the careful balance between privacy and accuracy.
In the non-federated setting, the level of noise must be adjusted to reasonably satisfy the definition of DP while ensuring that the raw data is not modified enough to affect the accuracy of any resulting analysis.
Following the new privacy definitions for FL, in the local privacy setting Li \emph{et al.}~\cite{lifederated} define privacy to distinguish between samples from a single device, instead of only between different devices.
Furthermore, a different Li \emph{et al.}~\cite{limeta} add a further stipulation to this sample-specific privacy to offer protection from all downstream agents, particularly the central server in a FL network.
The two contributions discussed above allow increased flexibility to improve privacy specifically in a federated setting.
To further improve the privacy-accuracy trade-off in the federated setting, DP can be combined with secure multi-party communication (SMC) to develop protocols that guarantee privacy without sacrificing accuracy~\cite{truex}.

Recent relevant works in the FL setting include Auxo~\cite{liu}, a systematic clustering mechanism that identifies clients with statistically similar data distributions, before adapting its training to achieve better model performance.
This addresses the problem of statistical heterogeneity (SH) identified in Section~\ref{sec:intro}, but in the training phase instead of the client selection phase.
Hu {et al.}~\cite{hu} improve the efficiency of such a heterogeneous model through the utilization of singular value decomposition.

Aledhari \emph{et al.}~\cite{aledhari} give a comprehensive overview of FL in terms of applicability.
Instead of focusing solely on the widely explored mobile keyboard prediction software, the authors describe how FL can be used in novel applications such as the ranking of browser history suggestions, medical imaging and augmented reality.
Such applications can be readily applied to the FA setting in a similar way to the word analytic and music recognizer discussed in Section~\ref{sec:intro}.

Wang \emph{et al.}~\cite{wang} introduce the first method that intelligently selects clients based on SH in an FA environment.
Their algorithm, named \emph{FedACS}, introduces a \emph{skewness tolerance parameter}, $\lambda$, so that a lower $\lambda$ indicates that the sampling process will be less tolerant towards clients with more unusual data, resulting in a more accurate analysis and therefore combating the problem of SH in FA.
In practice, the less typical clients are removed through a pairwise comparison method on an initially random sample of clients.
Therefore, the parameter $\lambda$ must not be too low, as this prevents sufficient raw data from being collected.

\section{Preliminaries} \label{sec:prelims}

First, we mathematically define the notion of differential privacy (DP). Recall from Section~\ref{sec:relwork} that DP requires that the contribution of any single user to a dataset does not have a significant impact on the outcome of its analysis.

\begin{defn} \label{defn:dp}
Consider two datasets $\vec{X}, \vec{X}' \in\mathbb{X}^{n}$ that differ only in one element, let $\varepsilon\geq 0$ and $\delta\in [0, 1]$. The randomized mechanism $\mathcal{M}:\mathbb{X}^{n} \rightarrow\mathbb{Y}$ is $(\varepsilon, \delta)$-differentially private if for all $E \subseteq\mathbb{Y}$:
\[
\mathbb{P}[\mathcal{M}(\vec{X})\in E] \leq e^{\varepsilon}\mathbb{P}[\mathcal{M}(\vec{X}')\in E] + \delta~\textnormal{\cite{balleprivacyblanket}}.
\]
\end{defn}

The formal definition of local differential privacy (LDP) is similar, except it considers changing a user's value $x$ to another value $x'$.
\newline
\newline
We now address the trade-off between the desired number of clients, their skewness, the privacy of the data and its computational cost. If each sample is smaller, then any formal differential privacy guarantee will be weaker, because the more sparse distributions are likely to be more varied, increasing the likelihood of individual data points being exposed.
However, if each sample is larger, then the computational cost of each round of FA is greater.
Mathematically, this trade-off between privacy and cost can be expressed in terms of the parameter $\varepsilon$ in the formal definition of LDP~\cite{balleprivacyblanket} and the number of clients $n$ in each sample.

As the first method that specifically combats SH in the FA setting, \emph{FedACS} has room for improvement.
In particular, its pairwise comparison and elimination method incurs a high computational cost, and can be improved greatly by replacing it with a single-client skewness measure.

A good initial attempt to tackle client skewness in an efficient way is \emph{FedCCEA}, a federated client contribution evaluation accuracy approximation proposed by Shyn \emph{et al.}~\cite{shyn}.
\emph{FedCCEA} measures the importance of each client from the feasibility of their initial data and uses this information to adjust the number of credits given to each client.
Although a specific measure for skewness is not given, this method has a reasonable calculation time regardless of the number of clients participating in the FL framework.

A more detailed approach is given by Wu \emph{et al.}~\cite{wufast}, whose \emph{FedAdp} algorithm incorporates clients whose local gradient is sufficiently correlated with the global gradient.
This direct measure of SH quantifies the contribution of each client during each global round based on angle $\theta_i(t)$, defined as:
\[
\theta_i(t) = \arccos \!\left( \frac{\langle \nabla F(\bm{w}(t)), \nabla F_i(\bm{w}(t)) \rangle}{\| \nabla F(\bm{w}(t)) \| \| \nabla F_i(\bm{w}(t)) \|} \right).
\]

Finally, a useful result from Dwork \emph{et al.}~\cite{dworkgauss} can be used to provide a shortcut to measuring the statistical heterogeneity of a dataset in matrix form.
The authors have shown that a large eigenvalue gap in this matrix representation, found using principal component analysis (PCA), improves the accuracy of a DP release of this dataset.
As a large eigenvalue gap in a matrix representation also implies a SH dataset, these two properties can be combined to allow the statistical heterogeneity of a dataset to be measured using PCA.
As this result scales well with a large dataset, it is a contender to be combined with \emph{FedACS}~\cite{wang} to formulate a differentially private FA framework that incorporates statistically heterogeneous datasets.

\subsection{Measures of statistical heterogeneity} \label{subsec:shmeasures}

To further the preliminary approaches given so far, the discussion now focuses on the problem of measuring the statistical heterogeneity of a $d$-dimensional vector $\vec{x}_{i} = (x_{i}^{(1)}, \dots, x_{i}^{(d)}) \in [0,1]^{d}$, which represents a contribution $\vec{x}_{i}$ of a user $i$.
It would be easiest to modify one or more existing technique(s), if it is possible, to solve this more specific problem.

A number of candidates were investigated, including basic dispersion~\cite{eliazarmax}, Gini and Pietra indices~\cite{eliazarpietra}, Shannon entropy~\cite{nunes} and statistics relating to meta-analysis~\cite{hoaglin}.
Following an initial comparison, the Pietra index and the statistic $I^{2}$~\cite{lin} were chosen as the two most promising approaches.
Both measure statistical heterogeneity in a more precise way than the rest, and are also useful in the real world~\cite{eliazarpietra, lin}.

After concluding that implementing the Pietra index would be more complex than for $I^{2}$, the latter was chosen as the goal, with dispersion as the starting point and the statistic $Q$~\cite{hoaglin} as an intermediate step.
Although dispersion has limited scope and applications, it is inextricably linked to $I^{2}$ via $Q$.

\emph{Dispersion} measures statistical heterogeneity through the fluctuations of the distribution of the values of each $d$-dimensional vector $\vec{x}_{i}$ of the dataset $\vec{X} = (\vec{x}_{1}, \dots, \vec{x}_{n})$ around its mean $\vec{\mu} = (\mu^{(1)}, \dots, \mu^{(d)}) \in [0,1]^{d}$.

\begin{defn} \label{defn:mu}
The mean $\vec{\mu}$ of the distribution of vectors can be written as $\vec{\mu} = \!\left( \sum_{i=1}^{n} \vec{x}_{i} \right) /n$.
\end{defn}

\begin{defn}
\label{defn:disp}
The dispersion $D$ of each vector $\vec{x}_{i}$ is given by the functional:
\[
D(\vec{x}_{i}) = \big|\hspace{0.2mm}\vec{x}_{i} - \vec{\mu}\hspace{0.2mm}\big|^{p},
\]

where the parameter $p \geq 1$ is called the \emph{dispersion exponent}~\cite{eliazarmax}.
\end{defn}

\begin{defn}
Using the characterization of the mean $\vec{\mu}$, the dispersion $D$ of the dataset $\vec{X}$ is given by $D(\vec{X}) = (\sum_{i=1}^{n} D(\vec{x}_{i}))/n$.
\end{defn}

While dispersion can detect variations in a vector $\vec{x}_{i}$ as a whole, it cannot account for variations between its coordinates.
$Q$, a statistic typically used in the meta-analysis of two studies, can detect inter-vector variations and normalize its output via the \emph{weighted mean} $\vec{\bar{\mu}}$.
Originating as `Cochran's $Q$ test' in 1954~\cite{cochran}, the initial version of $Q$ could detect the variation of a mathematical model via its estimate $x_{i}$, after which the normalization was to be computed manually.
In the decades since Cochran's paper, $Q$
has been developed into a identifier of heterogeneity, to conclude whether two studies produce similar effects or not.
The consensus in the literature~\cite{hoaglin} is that if $Q < 0.1$, statistical heterogeneity is present.
Therefore, $Q$ is only able to quantify statistical heterogeneity in a binary way.
The results in Section~\ref{subsec:q} enable progress towards the closely related statistic $I^{2}$, which can evaluate the extent of statistical heterogeneity.
Corresponding results for the statistic $I^{2}$ are in Section~\ref{subsec:isquared}.

\begin{defn} \label{defn:barmu}
The weighted mean $\vec{\bar{\mu}}$ of the vector $\vec{x}_{i}$ can be written as:
\[
\vec{\bar{\mu}} = \frac{\!\left(\sum_{i=1}^{n} w_{i} \vec{x}_{i}\right)/n}{\sum_{i=1}^{n} w_{i}},
\]

where $w_{i} = 1/s_{i}^{2}$, with $s_{i}^{2}$ the \emph{within-vector variance} of $\vec{x}_{i}$.
\end{defn}

It can be seen that when $s_{i}^{2}=1$, the weighted mean $\vec{\bar{\mu}}$ is equal to the mean $\vec{\mu}$.

\begin{defn}
The statistic $Q$ of each vector $\vec{x}_{i}$ is given by the weighted sum of the squared deviations:
\[
Q(\vec{x}_{i}) = w_{i}(\vec{x}_{i} - \pvec{\bar{\mu}})^{2}.
\]
\end{defn}

\begin{defn}
The statistic $Q$ of the dataset $\vec{X}$ is given by $Q(\vec{X}) = (\sum_{i=1}^{n} Q(\vec{x}_{i}))/n$.
\end{defn}

Our objective is to measure how much a private measure of dispersion, $Q$ or $I^2$ deviates from the corresponding true value.
We will use the \emph{mean squared error}, defined as the average squared difference, for this purpose.

\subsection{Analytic Gaussian Mechanism} \label{subsec:theagm}

The \emph{Analytic Gaussian Mechanism (AGM)} provides an exact formula for calibrating Gaussian noise depending on the $L_{2}$ sensitivity $\Delta$, and the privacy parameters $\varepsilon$ and $\delta$.
There are many advantages of using this mechanism, as opposed to the Classical Gaussian Mechanism (CGM), used to denote Gaussian sampling in Section~\ref{sec:theory}.

\begin{thm}\textnormal{(Classical Gaussian Mechanism (CGM).} \label{thm:cgm}
For any $\varepsilon$, $\delta \in (0, 1)$, the Gaussian output perturbation mechanism with $\Delta = \sqrt{d}/n$ and $\sigma = \Delta \sqrt{2 \log(1.25/\delta)}/\varepsilon$ is ($\varepsilon, \delta$)-differentially private~\textnormal{\cite{balleimproving}}.
\end{thm}

The AGM seeks the optimum instead of a feasible solution, reducing the amount of noise required for differentially private machine learning.
Applications include stochastic gradient descent, which is well suited for the vector data used in Section~\ref{sec:theory}.
Additionally, the AGM is applicable for arbitrary $\varepsilon$, as opposed to $\varepsilon < 1$ in the CGM.

Solving the AGM requires finding a solution to a simple optimization problem involving the \emph{cumulative distribution function} $\Phi(t)$.
An intermediate result provides a necessary and sufficient condition for differential privacy in terms of the privacy loss random variables $L_{M,x,x'}$ and $L_{M,x',x}$ of a vector-valued mechanism $M$ on a pair of neighboring inputs $x \simeq x'$.
It remains to specialize this result for a Gaussian output perturbation mechanism, in terms of the Gaussian cumulative distribution function $\Phi$, and to incorporate the global $L_{2}$ sensitivity $\Delta$.

\begin{thm}\textnormal{(Analytic Gaussian Mechanism).} \label{thm:main}
Let $f : \mathbb{X} \rightarrow \mathbb{R}^{d}$ be a function with global $L_{2}$ sensitivity $\Delta = \sqrt{d}/n$.
For any $\varepsilon \geq 0$ and $\delta \in [0,1]$, the Gaussian output perturbation mechanism $M(x) = f(x) + Z$ with $Z \sim \mathcal{N}(0, \sigma^{2} I)$ is $(\varepsilon, \delta)$-DP if and only if:
\[
\Phi \!\left( \frac{\Delta}{2\sigma} - \frac{\varepsilon\sigma}{\Delta} \right) - e^{\varepsilon} \Phi \!\left( -\frac{\Delta}{2\sigma} - \frac{\varepsilon\sigma}{\Delta} \right) \leq \delta~\textnormal{\cite{balleimproving}}.
\]
\end{thm}

There are a number of approaches for solving the inequality in Theorem~\ref{thm:main}.
An analytic expression for $\sigma$ could be found using the upper and lower bounds for the tail of the Gaussian cumulative distribution function, but these tail bounds are not tight in this non-asymptotic setting.
Instead, $\sigma$ could be found using a numerical algorithm.

\begin{fact} \label{fact:erf}
The Gaussian cumulative distribution function can be written as:
\[
\Phi(t) = \frac{\!\left( 1 + \erf \!\left( \frac{t}{\sqrt{2}} \right) \right)}{2},
\]

where $\erf$ is the standard error function.
\end{fact}

\begin{thm}
Let $f$ be a function with global $L_{2}$ sensitivity $\Delta$.
For any $\varepsilon > 0$ and $\delta \in (0, 1)$, the mechanism described in Algorithm~\ref{alg:theagm} is $(\varepsilon, \delta)$-DP~\textnormal{\cite{balleimproving}}.
\end{thm}

\begin{algorithm}[ht]
\DontPrintSemicolon
\SetArgSty{textnormal}
\KwPuI{$f$, $\Delta$, $\varepsilon$, $\delta$}\\
\KwPrI{$x$}\\
\medskip
Let $\delta_{0} = \Phi(0) - e^{\varepsilon}
\Phi(-\sqrt{2\varepsilon})$\\
\medskip
\eIf{$\delta \geq \delta_{0}$}
{Define $B_{\varepsilon}^{+}(v) = \Phi(\sqrt{\varepsilon v}) - e^{\varepsilon} \Phi (-\sqrt{\varepsilon(v+2)})$\\
Compute $v^{*} = \sup \{ v \in \mathbb{R}_{\geq 0} : B_{\varepsilon}^{+}(v) \leq \delta \}$\\
Let $\alpha = \sqrt{(1 + v^{*})/2} - \sqrt{v^{*}/2}$\\}
{Define $B_{\varepsilon}^{-}(u) = \Phi(-\sqrt{\varepsilon u}) - e^{\varepsilon} \Phi (-\sqrt{\varepsilon(u+2)})$\\
Compute $u^{*} = \inf \{ u \in \mathbb{R}_{\geq 0} : B_{\varepsilon}^{-}(u) \leq \delta \}$\\
Let $\alpha = \sqrt{(1 + u^{*})/2} + \sqrt{u^{*}/2}$\\}
\medskip
Let $\sigma = (\alpha\Delta)/\sqrt{2\varepsilon}$\\
Return $f(x) + \mathcal{N}(0, \sigma^{2} I)$
\caption{Analytic Gaussian Mechanism}
\label{alg:theagm}
\end{algorithm}

\bigskip

Using Fact~\ref{fact:erf} to compute $\Phi(t)$ based on the standard error function $\erf$, it is straightforward to implement a solver\footnote{A Python implementation of Algorithm~\ref{alg:theagm} can be found at \newline \url{https://github.com/BorjaBalle/analytic-gaussian-mechanism}.} for finding the values $v^{*}$ and $u^{*}$ required in Algorithm~\ref{alg:theagm}.
For example, Newton's method can be used to find the root $v^{*}$, using the fact that $B_{\varepsilon}^{+}(v)$ is monotonically increasing.
In practice, binary search can be used to find the smallest $k \in \mathbb{N}$ such that $B_{\varepsilon}^{+}(2^{k}) > \delta$.

\section{Theoretical Results} \label{sec:theory}

In this section we present the novel contributions of this work.
We develop formulae for private measures of statistical heterogeneity from first principles, and then evaluate their accuracy via mean squared error and confidence intervals.
We start with expressing noisy versions of the simplest measure, dispersion, building upon existing knowledge presented in Section~\ref{subsec:shmeasures}.

\subsection{Basic dispersion} \label{subsec:disp}

Although the mean $\vec{\mu}$ aggregates the vectors from all $n$ users, it is necessary to add noise from a distribution to protect it from difference attacks, and therefore satisfy the definition of differential privacy.
In this scenario, the most effective way to achieve approximate differential privacy is to add noise from a Gaussian distribution.
This is because when there are many coordinates to perturb at a time, the use of the $L_{2}$ norm in conjunction with the Gaussian distribution only requires the noise for each vector $\vec{x}_{i} = (x_{i}^{(1)}, \dots, x_{i}^{(d)}) \in [0,1]^{d}$ to be scaled by $\sqrt{d}$, instead of a scaling of $d$ for the $L_{1}$ norm in conjunction with the Laplace or exponential distribution~\cite{balleimproving}.

The Gaussian noise will be added in a distributed manner, which works well for this problem because the $n$ clients would most likely be separate.
Using the property that the composition of $n$ vectors, each with Gaussian noise, results in a private aggregated dataset with all of the Gaussian noise aggregated as well~\cite{dworkdng}, we predict that the accuracy of the upcoming results will not be affected by whether the noise is added in a centralized or a distributed manner.
The incorporation of a secure aggregation protocol from the federated analytics setting ensures that only an aggregate function of the vectors can be learned~\cite{kairouz}.
This distributed approach will also allow paths towards satisfying the definitions of the local and shuffle models of differential privacy~\cite{balleprivacyblanket}.

In what follows, we focus on the important case of $p=2$, which corresponds to the variance -- the most appropriate measure for statistical analysis.

\begin{lem} \label{lem:noisymu}
Using the assumption that the norm of the vector $\vec{x}_{i}$ is bounded by $\sqrt{d}/n$~\textnormal{\cite{balleimproving}}, we can produce a noisy mean $\pvec{\mu}'$ with $(\varepsilon_{1}, \delta_{1})$-differential privacy:
\[
\pvec{\mu}' = \vec{\mu} + Y_{1}, \quad \textit{where} \hspace{0.5em} Y_{1} \sim \mathcal{N}^{d}(\xi_{1}^{2}),
\]

and $\mathcal{N}^{d}(\xi_{1}^{2})$ denotes $d$-dimensional sampling from a Gaussian distribution with mean 0 and variance $\xi_{1}^{2} = \frac{2d \log(1.25/\delta_{1})}{n^{2} \varepsilon_{1}^{2}}$.
\end{lem}

\begin{proof}
By Theorem~\ref{thm:cgm}.
\end{proof}

\begin{lem} \label{lem:noisydisp}
In a similar way, noisy dispersions $D'$ and $D''$ of the dataset $\vec{X}$ can be produced with ($\varepsilon_{2},\delta_{2}$)-differential privacy:
\[
D'(\vec{X}) = \sum_{i=1}^{n} (\vec{x}_{i} - \pvec{\mu}')^{2} \ \ \enspace
and \quad D''(\vec{X}) = D'(\vec{X}) + Y_{2}, 
\]

where $Y_{2} \sim \mathcal{N}^{n\times d}(\xi_{2}^{2})$, and $\mathcal{N}^{n\times d}(\xi_{2}^{2})$ denotes $n\times d$-dimensional sampling from a Gaussian distribution with mean 0 and variance $\xi_{2}^{2} = \frac{2d \log(1.25/\delta_{2})}{n^{2} \varepsilon_{2}^{2}}$.
\end{lem}

\begin{proof}
By Theorem~\ref{thm:cgm}.
\end{proof}

\begin{thm} \label{thm:disp}
Combining Definition~\ref{defn:mu} with Lemmas~\ref{lem:noisymu} and \ref{lem:noisydisp}, an expression for the noisy dispersion $D''$ of the dataset $\vec{X}$, with ($\varepsilon = \varepsilon_{1} + \varepsilon_{2}, \delta = \delta_{1} + \delta_{2}$)-differential privacy, can be written as the following:
\[
D''(\vec{X}) = \sum_{i=1}^{n} \Big[ \!\left( ( \vec{x}_{i} - \pvec{\mu}) - Y_{1} \right)^{2} + Y_{2} \Big],
\]

where $Y_{1} \sim \mathcal{N}^{d}(\xi_{1}^{2})$, $Y_{2} \sim \mathcal{N}^{d}(\xi_{2}^{2})$, and $\mathcal{N}^{d}(\xi_{1}^{2})$, $\mathcal{N}^{d}(\xi_{2}^{2})$ denote $d$-dimensional sampling from Gaussian distributions with mean 0 and variances $\xi_{1}^{2} = \frac{2d \log(1.25/\delta_{1})}{n^{2} \varepsilon_{1}^{2}}$, $\xi_{2}^{2} = \frac{2d \log(1.25/\delta_{2})}{n^{2} \varepsilon_{2}^{2}}$ respectively.
\end{thm}

A key metric that is useful for the experimental evaluation of the expressions in Lemma~\ref{lem:noisymu}, Lemma~\ref{lem:noisydisp} and Theorem~\ref{thm:disp} is the \emph{mean squared error}, a measure of how much the noisy mean, dispersion or other statistic deviates from the corresponding true value without the noise.
Corollaries~\ref{cor:msemean} and \ref{cor:msedisp} display formulae for the mean squared errors of the noisy mean and dispersion respectively.

\begin{cor} \label{cor:msemean}
Rearranging Lemma~\ref{lem:noisymu}, an expression for the mean squared error of $\pvec{\mu}'$, with ($\varepsilon_{1}, \delta_{1}$)-differential privacy, can be written as:
\[
\textnormal{MSE}\hspace{0.1em}(\pvec{\mu}') = (\pvec{\mu}' - \pvec{\mu})^{2} = Y_{1}^{2},
\]

where $Y_{1} \sim \mathcal{N}^{d}(\xi_{1}^{2})$.
\end{cor}

\begin{restatable}{cor}{primedispmse} \label{cor:msedisp}
Rearranging Lemma~\ref{lem:noisydisp} and Theorem~\ref{thm:disp}, an expression for the mean squared error of $D''$, with ($\varepsilon, \delta$)- differential privacy, can be written as the following:
\[
\textnormal{MSE}\hspace{0.1em}(D''(\vec{X})) = \left( \sum_{i=1}^{n} \!\left[ Y_{1} \big( Y_{1} - 2(\vec{x}_{i} - \pvec{\mu}) \big) + Y_{2} \right]^{2} \right) /n,
\]

where $Y_{1} \sim \mathcal{N}^{d}(\xi_{1}^{2})$ and $Y_{2} \sim \mathcal{N}^{d}(\xi_{2}^{2})$.
\end{restatable}

\begin{proof}
See Appendix~\ref{app:mse}.
\end{proof}

It only remains now to give a confidence interval for dispersion in terms of the parameters $n$ and $\varepsilon$.
Lemma~\ref{lem:dispest} gives the average squared error of the difference between each vector $\vec{x}_{i}$ and the true and approximate means $\vec{\mu}$ and $\pvec{\mu}'$ respectively.

\begin{restatable}{lem}{primedispest} \label{lem:dispest}
Using the definitions of $\vec{\mu}$ and $\pvec{\mu}'$, the variance of $D''$ (the estimated average dispersion) can be written as:
\[
\textnormal{Var}\hspace{0.1em}(D''(\vec{X})) = \!\left[ \Big( \sum_{i=1}^{n} (\vec{x}_{i} - \pvec{\mu})^{2} \Big)/n - \Big( \sum_{i=1}^{n} (\vec{x}_{i} - \pvec{\mu}')^{2} \Big)/n \right]^{2} \hspace{-0.1em} = Y_{1}^{\hspace{0.05em} 4},
\]

where $Y_{1} \sim \mathcal{N}^{d}(\xi_{1}^{2})$.
\end{restatable}

\begin{proof}
See Appendix~\ref{app:est}.
\end{proof}

Theorem~\ref{thm:dispconf} computes a 95\% confidence interval for dispersion using the $z$-score of 1.96 and the property that for the random variable $Y_{1} \sim \mathcal{N}^{d}(\xi_{1}^{2})$, $\mathbb{E}[Y_{1}^{2n}] = (2n - 1)!!\hspace{0.1em}\xi_{1}^{2n}$~\cite{papoulis}.

\begin{restatable}{thm}{primedispconf} \label{thm:dispconf}
A 95\% confidence interval for $D$ (the true average dispersion) is given by:
\[
D(\vec{X}) = D''(\vec{X}) \pm \frac{7.84 \sqrt{6} \hspace{0.1em} (\xi_{1}^{2})^{2}}{\sqrt{n}}, \quad \textit{where} \hspace{0.5em} \xi_{1}^{2} = \frac{2d \log(1.25/\delta_{1})}{n^{2} \varepsilon_{1}^{2}} \hspace{0.1em}. \label{eq:dispconf}
\]
\end{restatable}

\begin{proof}
See Appendix~\ref{app:conf}.
\end{proof}

Note that the confidence interval given in Theorem~\ref{thm:dispconf} scales with $1/n^{4.5}$, showing that the accuracy of the true average dispersion improves as $n$ increases.

\subsection{\texorpdfstring{$Q$}{Q} in meta-analysis} \label{subsec:q}

To satisfy the definition of differential privacy for $Q$, the same steps are to be taken as for dispersion, except the weighted mean $\vec{\bar{\mu}}$ is used instead of the mean $\vec{\mu}$.

\begin{lem} \label{lem:noisybarmu}
Using the assumption that the norm of the vector $\vec{\bar{\mu}}$ is bounded by 1, we can produce a noisy weighted mean $\pvec{\bar{\mu}}'$ with ($\varepsilon_{1}, \delta_{1}$)-differential privacy:
\[
\pvec{\bar{\mu}}' = \vec{\bar{\mu}} + Z_{1}, \quad \textit{where} \hspace{0.5em} Z_{1} \sim \mathcal{N}^{d}(\eta_{1}^{2}),
\]

and $\mathcal{N}^{d}(\eta_{1}^{2})$ denotes $d$-dimensional sampling from a Gaussian distribution with mean 0 and variance $\eta_{1}^{2} = \frac{2d \log(1.25/\delta_{1})}{n^{2} \varepsilon_{1}^{2}} \hspace{0.1em}$.
\end{lem}

\begin{proof}
By Theorem~\ref{thm:cgm}.
\end{proof}

\begin{lem} \label{lem:noisyq}
In a similar way, noisy $Q'$ and $Q''$ of the dataset $\vec{X}$ can be produced with ($\varepsilon_{2}, \delta_{2}$)-differential privacy:
\[
Q'(\vec{X}) = \sum_{i=1}^{n} w_{i} (\vec{x}_{i} - \pvec{\bar{\mu}}')^{2} \ \ \enspace and \quad Q''(\vec{X}) = Q'(\vec{X}) + Z_{2}, 
\]

where $Z_{2} \sim \mathcal{N}^{n\times d}(\eta_{2}^{2})$, and $\mathcal{N}^{n\times d}(\eta_{2}^{2})$ denotes $n\times d$-dimensional sampling from a Gaussian distribution with mean 0 and variance $\eta_{2}^{2} = \frac{2d \log(1.25/\delta_{2})}{n^{2} \varepsilon_{2}^{2}}$.
\end{lem}

\begin{proof}
By Theorem~\ref{thm:cgm}.
\end{proof}

\begin{thm} \label{thm:noisyq}
Combining Definition~\ref{defn:barmu} with Lemmas~\ref{lem:noisybarmu} and \ref{lem:noisyq}, an expression for a noisy $Q''$ of the dataset $\vec{X}$, with ($\varepsilon = \varepsilon_{1} + \varepsilon_{2}, \delta = \delta_{1} + \delta_{2}$)-differential privacy, can be written as the following:
\[
Q''(\vec{X}) = \sum_{i=1}^{n} \Big[(w_{i} (\vec{x}_{i} - \pvec{\bar{\mu}}) - Z_{1})^{2} + Z_{2} \Big],
\]

where $w_{i} = 1/s_{i}^{2}$ with $s_{i}^{2}$ the within-vector variance of $\vec{x}_{i}$, $Z_{1} \sim \mathcal{N}^{d}(\eta_{1}^{2})$, $Z_{2} \sim \mathcal{N}^{d}(\eta_{2}^{2})$, and $\mathcal{N}^{d}(\eta_{1}^{2})$, $\mathcal{N}^{d}(\eta_{2}^{2})$ denote $d$-dimensional sampling from Gaussian distributions with mean 0 and variances $\eta_{1}^{2} = \frac{2d \log(1.25/\delta_{1})}{n^{2} \varepsilon_{1}^{2}}$ and $\eta_{2}^{2} = \frac{2d \log(1.25/\delta_{2})}{n^{2} \varepsilon_{2}^{2}}$ respectively.
\end{thm}

It can be seen that the formulae in Theorems~\ref{thm:disp} and \ref{thm:noisyq} differ only in the weight term $w_{i}$ and specific formula for $\vec{\bar{\mu}}$ (which also features $w_{i}$).

\begin{alg} An algorithm to calculate $Q''$: \label{alg:noisyq}
\begin{itemize}
    \item Each user locally calculates the within-vector variance $s_{i}^{2}$ of their vector $\vec{x}_{i}$.
    \item Each user locally computes $w_{i} = 1/s_{i}^{2}$.
    \item Trusted server computes weighted mean
    \[
    \vec{\bar{\mu}} = \frac{\!\left(\sum_{i=1}^{n} w_{i} \vec{x}_{i}\right)/n}{\sum_{i=1}^{n} w_{i}}.
    \]
    \item Trusted server adds noise with privacy parameters $\varepsilon_{1}$ and $\delta_{1}$ to $\vec{\bar{\mu}}$ to get $\pvec{\bar{\mu}}'$.
    \item Trusted server computes $Q'$ using private weighted mean $\pvec{\bar{\mu}}'$: \newline $Q'(\vec{X}) = \sum_{i=1}^{n} w_{i}(\vec{x}_{i} - \pvec{\bar{\mu}}')^{2}$.
    \item Trusted server adds noise with privacy parameters $\varepsilon_{2}$ and $\delta_{2}$ to $Q'$ to get $Q''$.
\end{itemize}
\end{alg}

Corollaries~\ref{cor:msewmean} and \ref{cor:msenoisyq} display formulae for the mean squared errors of the noisy weighted mean and $Q''$ respectively.

\begin{cor} \label{cor:msewmean}
Rearranging Lemma~\ref{lem:noisybarmu}, an expression for the mean squared error of $\pvec{\bar{\mu}}'$, with ($\varepsilon_{1}, \delta_{1}$)-differential privacy, can be written as:
\[
\textnormal{MSE}\hspace{0.1em}(\pvec{\bar{\mu}}') = (\pvec{\bar{\mu}}' - \pvec{\bar{\mu}})^{2} = Z_{1}^{2},
\]

where $Z_{1} \sim \mathcal{N}^{d}(\eta_{1}^{2})$.
\end{cor}

\begin{cor} \label{cor:msenoisyq}
Rearranging Lemma~\ref{lem:noisyq} and Theorem~\ref{thm:noisyq}, an expression for the mean squared error of the noisy statistic $Q''$, with ($\varepsilon, \delta$)-differential privacy, can be written as the following:
\[
\textnormal{MSE}\hspace{0.1em}(Q''(\vec{X})) = \left( \sum_{i=1}^{n} \!\left[ w_{i} \hspace{0.1em} Z_{1} \big( Z_{1} - 2(\vec{x}_{i} - \pvec{\bar{\mu}}) \big) + Z_{2} \right]^{2} \right) /n,
\]

where $Z_{1} \sim \mathcal{N}^{d}(\eta_{1}^{2})$ and $Z_{2} \sim \mathcal{N}^{d}(\eta_{2}^{2})$.
\end{cor}

\begin{proof}
See Appendix~\ref{app:mse}.
\end{proof}

The estimated average $Q$ is now given in Lemma~\ref{lem:qest}, in terms of the parameters $w_{i} = 1/s_{i}^{2}$, $n$ and $\varepsilon$.

\begin{lem} \label{lem:qest}
Using the definitions of $\vec{\bar{\mu}}$ and $\vec{\bar{\mu}}'$, the variance of $Q''$ (the estimated average $Q$) can be written as:
\begin{align*}
\textnormal{Var}\hspace{0.1em}(Q''(\vec{X})) &= \!\left[ \Big( \sum_{i=1}^{n} w_{i} (\vec{x}_{i} - \pvec{\bar{\mu}})^{2} \Big)/n - \Big( \sum_{i=1}^{n} w_{i} (\vec{x}_{i} - \pvec{\bar{\mu}}')^{2} \Big)/n \right]^{2} \hspace{-0.1em} \\
&= \sum_{i=1}^{n} w_{i} \hspace{0.05em} Z_{1}^{\hspace{0.05em} 4},
\end{align*}

where $Z_{1} \sim \mathcal{N}^{d}(\eta_{1}^{2})$.
\end{lem}

\begin{proof}
See Appendix~\ref{app:est}.
\end{proof}

Theorem~\ref{thm:qconf} computes a 95\% confidence interval for $Q$ using the $z$-score of 1.96 and the property that for the random variable $Z_{1} \sim \mathcal{N}^{d}(\eta_{1}^{2})$, $\mathbb{E}[Z_{1}^{2n}] = (2n - 1)!!\hspace{0.1em}\eta_{1}^{2n}$~\cite{papoulis}.

\begin{thm} \label{thm:qconf}
A 95\% confidence interval for (the true average) $Q$ is given by:
\[
Q(\vec{X}) = Q''(\vec{X}) \pm \sum_{i=1}^{n} \frac{7.84 w_{i} \sqrt{6} \hspace{0.1em} (\eta_{1}^{2})^{2}}{\sqrt{n}}, \quad \label{eq:qconf}
\]

where $\eta_{1}^{2} = \frac{2d \log(1.25/\delta_{1})}{n^{2} \varepsilon_{1}^{2}}$.
\end{thm}

\begin{proof}
See Appendix~\ref{app:conf}.
\end{proof}

In a similar way to dispersion (Theorem~\ref{thm:dispconf}), the confidence interval given in Theorem~\ref{thm:qconf} scales with $1/n^{4.5}$, showing that the accuracy of $Q$ improves as $n$ increases.

\subsection{\texorpdfstring{$I^{2}$}{I2} in meta-analysis} \label{subsec:isquared}

$I^{2}$ is a well-known statistic used in meta-analysis, an alteration of $Q$ that is able to quantify the extent of statistical heterogeneity in a non-binary way.

\begin{defn} \label{defn:isquared}
The statistic $I^{2}$ of the dataset $\vec{X}$ has expression
\[
I^{2}(\vec{X}) = \max \!\left\{ 0, \ 1 - \frac{n - 1}{Q(\vec{X})} \right\},
\]
\end{defn}

where $Q(\vec{X}) = \sum_{i=1}^{n} w_{i} (\vec{x}_{i} - \pvec{\bar{\mu}})^{2}$.
Note that if $Q < n - 1$, we assume $Q = n - 1$ so that $I^{2} \in [0, 1]$.
$I^{2}$ is often represented as a percentage, with $25\%$, $50\%$ and $75\%$ representing typical cut-off points to differentiate the extent of statistical heterogeneity~\cite{lin}.

\begin{thm} \label{thm:noisyisquared}
Noisy $(I^{2})''$ and $(I^{2})'''$ can be produced with ($\varepsilon_{3}, \delta_{3}$)-differential privacy:
\begin{align*}
(I^{2})''(\vec{X}) &= \max \!\left\{ 0, \ 1 - \frac{n - 1}{Q''(\vec{X})} \right\} \\
and \quad (I^{2})'''(\vec{X}) &= (I^{2})''(\vec{X}) + Z_{3},
\end{align*}

where $Z_{3} \sim \mathcal{N}^{n\times d}(\eta_{3}^{2})$, and $\mathcal{N}^{n\times d}(\eta_{3}^{2})$ denotes $n\times d$-dimensional sampling from a Gaussian distribution with mean 0 and variance $\eta_{3}^{2} = \frac{2d \log(1.25/\delta_{3})}{n^{2} \varepsilon_{3}^{2}}$.
\end{thm}

\begin{cor} \label{cor:msenoisyisquared}
Using Theorem~\ref{thm:noisyisquared}, an expression for the mean squared error of the noisy statistic $(I^{2})'''$, with ($\varepsilon, \delta$)-differential privacy, can be written as the following:
\[
\textnormal{MSE}\hspace{0.1em}((I^{2})'''(\vec{X})) = \frac{1}{n} \Bigg[ Z_{3} - \frac{n - 1}{Q''(\vec{X})} + \frac{n - 1}{Q(\vec{X})} \Bigg]^{2}.
\]
\end{cor}

\begin{restatable}{thm}{primeisquaredconf} \label{thm:isquaredconf}
A 95\% confidence interval for (the true average) $I^{2}$ is given by:
\[
I^{2}(\vec{X}) = (I^{2})'''(\vec{X}) \hspace{0.2em} \pm \sum_{i=1}^{n} \frac{0.625(n-1)}{w_{i}\sqrt{n}(\eta_{1}^{2})^{2}} \hspace{0.1em},
\]

where $w_{i} = 1/s_{i}^{2}$ with $s_{i}^{2}$ the within-vector variance of each $\vec{x}_{i}$, and $\eta_{1}^{2} = \frac{2d \log(1.25/\delta_{1})}{n^{2} \varepsilon_{1}^{2}}$.
\end{restatable}

\begin{proof}
See Appendix~\ref{app:conf}.
\end{proof}

\section{Experiments} \label{sec:expts}

The following extended experimental section has multiple purposes.
Primarily, it aims to consolidate the theoretical results in Section~\ref{sec:theory}, and select the mechanism that optimizes their accuracy.
Important questions to be answered include whether there is an advantage of using the Analytic Gaussian Mechanism (AGM) from Section~\ref{subsec:theagm} rather than the Classical Gaussian Mechanism (CGM), the cost of privacy on the accuracy of the AGM, and the cost of a distributed setting as opposed to a centralized setting.
Given a distributed setting, how robust is the AGM to different levels of statistical heterogeneity (SH)?

To answer each of these questions, we implement the AGM and CGM in Python.
We study the CIFAR datasets \cite{krizhevsky}, and the Fashion-MNIST dataset \cite{xiao} (a variant of the MNIST dataset \cite{lecun}).
The purpose of selecting multiple real-world datasets is to demonstrate the robustness of the AGM on datasets containing different data types and of various dimensions.
Because all of the datasets are image-based, it is necessary to reshape the raw data so that each of the two-dimensional images is represented by a one-dimensional vector containing the same number of coordinates as there were pixels in the image.
To keep privacy constant when it is not investigated, we fix its parameters $\varepsilon = 0.25$ and $\delta = 0.1$ unless otherwise specified.

\subsection{Statistics and Ratios} \label{subsec:stats}

In the below experiments $2\%$ of the images from each dataset are sampled such that there are equal numbers of each of the $10$ labels.
For the purposes of these experiments, the $20$ coarse labels of the CIFAR-100 dataset are combined into $10$ buckets consisting of labels $0$ and $1$, $2$ and $3$, etc.

\subsubsection*{Consolidating theoretical results}

We verify the validity of the main theoretical results in Section~\ref{sec:theory}.
We define the \emph{empirical mean squared error (EMSE)} of dispersion, $Q$ or $I^{2}$ to be the mean squared error, computed in the Python implementation, of the private expressions in Theorems~\ref{thm:disp}, \ref{thm:noisyq} or \ref{thm:noisyisquared} respectively.
We define the \emph{theoretical mean squared error (TMSE)} of dispersion, $Q$ or $I^{2}$ to be the output of Corollaries~\ref{cor:msedisp}, \ref{cor:msenoisyq} and \ref{cor:msenoisyisquared} respectively.
It would be a strong result for the EMSE and TMSE to be close to each other, and not necessarily expected.

\subsubsection*{AGM vs CGM}

Recall from Section~\ref{subsec:theagm} that one of the theoretical justifications of using the AGM is that it finds the optimal setting of the noise parameter $\sigma$, instead of simply a feasible solution.
This `working' solution is represented by the CGM, which is the mechanism by which the theorems in Section~\ref{sec:theory} are made private through the basic addition of Gaussian noise.
Therefore, we expect the AGM to have a significantly lower EMSE and TMSE than the CGM, with all other parameters fixed.

\subsubsection*{Cost of distributed setting}

Recall from Section~\ref{subsec:disp} that the Gaussian noise is added in a distributed manner.
In particular, each of the clients add their own noise before their $d$-dimensional vectors are brought together via a secure aggregation protocol~\cite{kairouz}.
It has been shown in the literature~\cite{dworkdng} that the accuracy of the results in Section~\ref{sec:theory} is not affected by whether the noise is added in a centralized or distributed manner.
Therefore, we are going to check this theory by comparing our distributed setting to a centralized setting, where the noise is added after the secure aggregation protocol.
We define the \emph{centralized mean squared error (CMSE)} to be the mean squared error of the Python implementation in the centralized setting.
Note that the CMSE equals the square of the noise term added after the secure aggregation protocol, and therefore it does not change whether the dispersion, $Q$ or $I^{2}$ is analyzed.
The empirical and theoretical versions are identical, so the CMSE can represent either of them.

\subsubsection*{Analysis}

First, we compare the EMSE and TMSE of the dispersion, $Q$ and $I^{2}$ metrics in Tables~\ref{tab:disp}, \ref{tab:q} and \ref{tab:isquared}.
Looking at the top two values of all `AGM' and `CGM' columns, the EMSE appears to be (almost) identical to the TMSE in most cases.
Table~\ref{tab:ratios}(a) displays the ratio EMSE/TMSE for all settings, confirming that this ratio is indeed (almost) equal to $1$ in all cases except for those involving the statistic $Q$.
Recall from Section~\ref{subsec:disp} that $Q$ is only able to quantify statistical heterogeneity in a binary way, unlike dispersion and $I^{2}$.
Therefore, we are able to verify the main theorems and corollaries in Section~\ref{sec:theory} in most cases, with a plausible explanation for the exceptions.

\begin{table}[hb]
\footnotesize
\begin{tabular}{lllll}\toprule
& \multicolumn{4}{c}{CIFAR-10} \\\cmidrule(lr){2-5}
& AGM & CGM & SD AGM & SD CGM \\
EMSE & \num{4.6654e-08} & \num{2.1043e-07} & \num{1.5915e-09} & \num{8.4053e-09} \\
TMSE & \num{4.6654e-08} & \num{2.1042e-07} & \num{1.5916e-09} & \num{8.4087e-09} \\
CMSE & \num{9.9906e-07} & \num{7.3631e-06} & \num{1.4178e-06} & \num{1.0349e-05} \\\midrule
& \multicolumn{4}{c}{CIFAR-100} \\
\cmidrule(lr){2-5}
& AGM & CGM & SD AGM & SD CGM \\
EMSE & \num{4.6806e-08} & \num{2.1779e-07} & \num{1.7159e-09} & \num{1.3884e-08} \\
TMSE & \num{4.6806e-08} & \num{2.1779e-07} & \num{1.7157e-09} & \num{1.3891e-08} \\
CMSE & \num{5.3006e-06} & \num{1.0585e-05} & \num{6.4108e-06} & \num{1.5455e-05} \\\midrule
& \multicolumn{4}{c}{Fashion-MNIST} \\
\cmidrule(lr){2-5}
& AGM & CGM & SD AGM & SD CGM \\
EMSE & \num{1.4472e-09} & \num{6.7639e-09} & \num{5.9609e-11} & \num{3.2919e-10} \\
TMSE & \num{1.4472e-09} & \num{6.7639e-09} & \num{5.9609e-11} & \num{3.2919e-10} \\
CMSE & \num{3.8181e-08} & \num{2.1295e-07} & \num{3.3557e-08} & \num{1.4850e-07} \\\bottomrule
\end{tabular}
\vspace{1mm}
\caption{Statistics of the dispersion metric for the AGM and CGM algorithms, including the EMSE, TMSE, CMSE and their standard deviations, using the CIFAR-10, CIFAR-100 and Fashion-MNIST datasets.\label{tab:disp}}
\end{table}

\begin{table}[p]
\footnotesize
\begin{tabular}{lllll}\toprule
& \multicolumn{4}{c}{CIFAR-10} \\\cmidrule(lr){2-5}
& AGM & CGM & SD AGM & SD CGM \\
EMSE & \num{5.657e-08} & \num{2.398e-07} & \num{1.365e-08} & \num{3.532e-08} \\
TMSE & \num{9.662e-06} & \num{1.479e-04} & \num{7.203e-05} & \num{2.757e-04} \\
CMSE & \num{9.991e-07} & \num{7.363e-06} & \num{1.418e-06} & \num{1.035e-05} \\\midrule
& \multicolumn{4}{c}{CIFAR-100} \\
\cmidrule(lr){2-5}
& AGM & CGM & SD AGM & SD CGM \\
EMSE & \num{4.984e-08} & \num{2.307e-07} & \num{5.038e-09} & \num{2.142e-08} \\
TMSE & \num{1.570e-05} & \num{5.006e-05} & \num{1.002e-05} & \num{6.478e-05} \\
CMSE & \num{5.301e-06} & \num{1.059e-05} & \num{6.411e-06} & \num{1.546e-05} \\\midrule
& \multicolumn{4}{c}{Fashion-MNIST} \\
\cmidrule(lr){2-5}
& AGM & CGM & SD AGM & SD CGM \\
EMSE & \num{1.484e-09} & \num{6.879e-09} & \num{7.388e-11} & \num{3.017e-10} \\
TMSE & \num{2.133e-08} & \num{1.765e-07} & \num{2.253e-08} & \num{1.663e-07} \\
CMSE & \num{3.818e-08} & \num{2.129e-07} & \num{3.356e-08} & \num{1.485e-07} \\\bottomrule
\end{tabular}
\vspace{1mm}
\caption{Statistics of the $Q$ metric for the AGM and CGM algorithms, including the EMSE, TMSE, CMSE and their standard deviations, using the CIFAR-10, CIFAR-100 and Fashion-MNIST datasets.\label{tab:q}}
\end{table}

\begin{table}[p]
\footnotesize
\begin{tabular}{lllll}\toprule
& \multicolumn{4}{c}{CIFAR-10} \\\cmidrule(lr){2-5}
& AGM & CGM & SD AGM & SD CGM \\
EMSE & \num{1.323e-08} & \num{4.013e-08} & \num{1.845e-08} & \num{4.818e-08} \\
TMSE & \num{1.336e-08} & \num{3.880e-08} & \num{1.801e-08} & \num{4.613e-08} \\
CMSE & \num{9.991e-07} & \num{7.363e-06} & \num{1.418e-06} & \num{1.035e-05} \\\midrule
& \multicolumn{4}{c}{CIFAR-100} \\
\cmidrule(lr){2-5}
& AGM & CGM & SD AGM & SD CGM \\
EMSE & \num{5.537e-09} & \num{2.466e-08} & \num{7.236e-09} & \num{2.719e-08} \\
TMSE & \num{6.011e-09} & \num{2.468e-08} & \num{7.464e-09} & \num{2.630e-08} \\
CMSE & \num{5.301e-06} & \num{1.059e-05} & \num{6.411e-06} & \num{1.546e-05} \\\midrule
& \multicolumn{4}{c}{Fashion-MNIST} \\
\cmidrule(lr){2-5}
& AGM & CGM & SD AGM & SD CGM \\
EMSE & \num{1.010e-07} & \num{3.235e-07} & \num{6.812e-08} & \num{3.514e-07} \\
TMSE & \num{1.008e-07} & \num{3.247e-07} & \num{6.845e-08} & \num{3.509e-07} \\
CMSE & \num{3.818e-08} & \num{2.129e-07} & \num{3.356e-08} & \num{1.485e-07} \\\bottomrule
\end{tabular}
\vspace{1mm}
\caption{Statistics of the $I^{2}$ metric for the AGM and CGM algorithms, including the EMSE, TMSE, CMSE and their standard deviations, using the CIFAR-10, CIFAR-100 and Fashion-MNIST datasets.\label{tab:isquared}}
\end{table}

\begin{table}[p]
\small
\begin{tabular}{llll}\toprule
& Dispersion & $Q$ & $I^{2}$ \\
CIFAR-10 & $0.164$ & $3.954$ & $251$ \\
CIFAR-100 & $3.954$ & $3.630$ & $274$ \\
Fashion-MNIST & $251$ & $0.911$ & $1314$ \\\bottomrule
\end{tabular}
\vspace{1mm}
\caption{Minimum values for the dispersion, $Q$ and $I^{2}$ metrics, using the CIFAR-10, CIFAR-100 and Fashion-MNIST datasets.\label{tab:minvals}}
\end{table}

\begin{table}[ht]
\small
\subfloat[Ratio EMSE/TMSE]{\scalebox{0.99}{
\begin{tabular}{llll}\toprule
& \multicolumn{3}{c}{CIFAR-10} \\\cmidrule(lr){2-4}
& Dispersion & $Q$ & $I^{2}$ \\
AGM & $1.0000$ & $0.0059$ & $0.9904$ \\
CGM & $1.0000$ & $0.0016$ & $1.0344$ \\\midrule
& \multicolumn{3}{c}{CIFAR-100} \\
\cmidrule(lr){2-4}
& Dispersion & $Q$ & $I^{2}$ \\
AGM & $1.0000$ & $0.0032$ & $0.9212$ \\
CGM & $1.0000$ & $0.0046$ & $0.9992$ \\\midrule
& \multicolumn{3}{c}{Fashion-MNIST} \\
\cmidrule(lr){2-4}
& Dispersion & $Q$ & $I^{2}$ \\
AGM & $1.0000$ & $0.0696$ & $1.0023$ \\
CGM & $1.0000$ & $0.0390$ & $0.9965$ \\\bottomrule
\end{tabular}}}
\qquad
\subfloat[Ratio AGM/CGM]{\scalebox{0.99}{
\begin{tabular}{llll}\toprule
& \multicolumn{3}{c}{CIFAR-10} \\\cmidrule(lr){2-4}
& Dispersion & $Q$ & $I^{2}$ \\
EMSE & $0.2217$ & $0.2359$ & $0.3296$ \\
TMSE & $0.2217$ & $0.0653$ & $0.3443$ \\
CMSE & $0.1357$ & $0.1357$ & $0.1357$ \\\midrule
& \multicolumn{3}{c}{CIFAR-100} \\
\cmidrule(lr){2-4}
& Dispersion & $Q$ & $I^{2}$ \\
EMSE & $0.2149$ & $0.2160$ & $0.2245$ \\
TMSE & $0.2149$ & $0.3136$ & $0.2435$ \\
CMSE & $0.5008$ & $0.5008$ & $0.5008$ \\\midrule
& \multicolumn{3}{c}{Fashion-MNIST} \\
\cmidrule(lr){2-4}
& Dispersion & $Q$ & $I^{2}$ \\
EMSE & $0.2140$ & $0.2157$ & $0.3123$ \\
TMSE & $0.2140$ & $0.1208$ & $0.2435$ \\
CMSE & $0.1793$ & $0.1793$ & $0.1793$ \\\bottomrule
\end{tabular}}}
\vspace{1mm}
\caption{Ratio EMSE/TMSE for AGM and CGM algorithms; ratio AGM/CGM for computation of EMSE, TMSE and CMSE.\label{tab:ratios}}
\end{table}

Next, we compare the AGM and CGM in Tables~\ref{tab:disp} and \ref{tab:q}. The columns `SD AGM' and `SD CGM' display the standard deviation of the AGM and CGM computations.
It is clear that the AGM always produces a lower or equal MSE than the CGM, whether the EMSE, TMSE or CMSE is used.
Table~\ref{tab:ratios}(b) confirms that the AGM is typically $0.1$-$0.3$ of the CGM for all settings.
Our results therefore confirm the theory in Section~\ref{subsec:theagm}, that the AGM produces more accurate results than the CGM.

Finally, we compare the EMSE and TMSE (distributed MSE) with the CMSE (centralized MSE).
It is clear in Tables~\ref{tab:disp}, \ref{tab:q} and \ref{tab:isquared}  that the CMSE is always the same or a similar order of magnitude compared to the EMSE and TMSE, whether the AGM or CGM is used.
This consolidates our hypothesis that the accuracy is not affected by whether the noise is added in a centralized or distributed way.
On the whole, it is slightly preferable to add Gaussian noise in a distributed manner.

We add context to the above results by presenting the minimum values for the dispersion, $Q$ and $I^{2}$ metrics in Table~\ref{tab:minvals}.
All values of EMSE, TMSE and CMSE in Tables~\ref{tab:disp}, \ref{tab:q} and \ref{tab:isquared} are many orders of magnitude smaller than the smallest true values of dispersion, $Q$ and $I^{2}$.
Therefore, these errors are small enough compared to the magnitude of the quantities in question.

\subsection{Statistical Heterogeneity and Privacy} \label{subsec:shprivacy}

Following the analysis in Section~\ref{subsec:stats}, we now focus on the EMSE as we vary the SH of the test data.

\subsubsection*{Different levels of heterogeneity}

Next, we will test the ability of the AGM to produce good accuracy whether a statistically heterogeneous dataset is used or not.
$2\%$ of the images are sampled from each dataset such that the following ratios are satisfied.
Note that the first case has been explored already in Section~\ref{subsec:stats}.

\begin{itemize}
    \item Equal numbers of each of 10 labels [1:1:1:1:1:1:1:1:1:1]
    \item Unequal numbers of each of 10 labels [91:1:1:1:1:1:1:1:1:1]
    \item Equal numbers of each of 5 labels [1:1:1:1:1]
    \item Unequal numbers of each of 5 labels [96:1:1:1:1]
    \item Equal numbers of each of 2 labels [1:1]
    \item Unequal numbers of each of 2 labels [99:1]
\end{itemize}

\subsubsection*{Cost of privacy}

We also explore the effect of adjusting the multiplicative privacy parameter $\varepsilon \in [0.25, 5]$ of the Gaussian noise added in a distributed manner.
When $\varepsilon$ is low, the privacy level is high because more noise is added.
The additive privacy parameter $\delta = 10^{-5}$ combines with $\varepsilon$ to satisfy the definition of differential privacy.

\begin{table}[ht]
\small
\subfloat[Non-SH sample vs SH sample]{\scalebox{0.9}{
\begin{tabular}{llll}\toprule
& \multicolumn{3}{c}{CIFAR-10} \\\cmidrule(lr){2-4}
& Dispersion & $Q$ & $I^{2}$ \\
10 labels & $+0.14$ & --$0.98$ & --$32.3$ \\
5 labels & $+0.00$ & --$1.46$ & --$37.0$ \\
2 labels & $+0.12$ & $+0.24$ & --$24.6$ \\\midrule
& \multicolumn{3}{c}{CIFAR-100} \\
\cmidrule(lr){2-4}
& Dispersion & $Q$ & $I^{2}$ \\
10 labels & --$0.59$ & $+0.35$ & --$5.99$ \\
5 labels & --$0.19$ & --$1.00$ & --$44.1$ \\
2 labels & $+0.57$ & $+0.66$ & --$24.3$ \\\midrule
& \multicolumn{3}{c}{Fashion-MNIST} \\
\cmidrule(lr){2-4}
& Dispersion & $Q$ & $I^{2}$ \\
10 labels & $+0.08$ & --$1.10$ & --$69.4$ \\
5 labels & $+0.03$ & --$0.65$ & --$96.0$ \\
2 labels & --$0.55$ & --$0.72$ & --$47.4$ \\\bottomrule
\end{tabular}}}
\qquad
\subfloat[Different numbers of labels (10v5 and 5v2)]{\scalebox{0.9}{
\begin{tabular}{llll}\toprule
& \multicolumn{3}{c}{CIFAR-10} \\\cmidrule(lr){2-4}
& Dispersion & $Q$ & $I^{2}$ \\
SH: 10v5 & --$0.34$ & --$0.04$ & --$20.5$ \\
SH: 5v2 & --$0.07$ & $+0.10$ & --$28.4$ \\
Non-SH: 10v5 & --$0.46$ & --$1.72$ & --$23.6$ \\
Non-SH: 5v2 & $+0.07$ & $+0.61$ & --$25.1$ \\\midrule
& \multicolumn{3}{c}{CIFAR-100} \\
\cmidrule(lr){2-4}
& Dispersion & $Q$ & $I^{2}$ \\
SH: 10v5 & $+0.36$ & $+0.09$ & --$35.1$ \\
SH: 5v2 & $+0.40$ & --$0.35$ & --$31.1$ \\
Non-SH: 10v5 & --$0.41$ & --$1.59$ & --$46.2$ \\
Non-SH: 5v2 & $+0.03$ & $+1.02$ & $+2.90$ \\\midrule
& \multicolumn{3}{c}{Fashion-MNIST} \\
\cmidrule(lr){2-4}
& Dispersion & $Q$ & $I^{2}$ \\
SH: 10v5 & --$0.04$ & $+0.47$ & --$15.3$ \\
SH: 5v2 & --$0.29$ & --$0.30$ & --$29.7$ \\
Non-SH: 10v5 & $+0.54$ & $+0.52$ & --$39.5$ \\
Non-SH: 10v5 & --$0.24$ & --$0.77$ & --$45.5$ \\\bottomrule
\end{tabular}}}
\vspace{1mm}
\caption{Comparing different levels of heterogeneity via percentage change of EMSE, averaged over all $\varepsilon \in [0.25, 5]$.\label{tab:changemse}}
\end{table}

\subsubsection*{Analysis}

The percentage changes in Table~\ref{tab:changemse}(a) are the result of computing the EMSE of a sample with equal numbers of each label (non-statistically heterogeneous), and then comparing it to a sample with unequal numbers of each label (statistically heterogeneous).
The number of labels, and their identity, stay constant in each comparison.
For dispersion and $Q$, the percentage change is small, always below $1.5\%$, and is positive half of the time and negative half of the time.
Therefore, for these metrics the AGM does not lose significant accuracy when a statistically heterogeneous dataset is used, regardless of the number of labels used.
For $I^{2}$, the percentage change is much larger, over $30\%$ on average, and almost always negative.
This implies that a statistically heterogeneous dataset causes the EMSE to decrease significantly, improving the accuracy of the AGM.
Therefore, the AGM can produce good accuracy whether a statistically heterogeneous dataset is used or not, and even better accuracy in the case of $I^{2}$.
Recall Table~\ref{tab:minvals} showed that the AGM EMSE for $I^{2}$ is sufficiently smaller than the minimum values of $I^{2}$ that we are handling.

Table~\ref{tab:changemse}(b) compares the percentage change in the EMSE of the AGM, when the number of labels is decreased and the statistical heterogeneity is kept constant.
The trends are similar to those in Table~\ref{tab:changemse}(a), with dispersion and $Q$ producing small positive and negative percentage changes and $I^2$ producing large negative percentage changes.
Therefore, the AGM can produce good accuracy regardless of the number of labels used, and even better accuracy when $I^2$ is used alongside a small number of labels.

\begin{figure*}[ht]
\centering
\subfloat[Dispersion for CIFAR-10]{\includegraphics[width=0.31\linewidth]{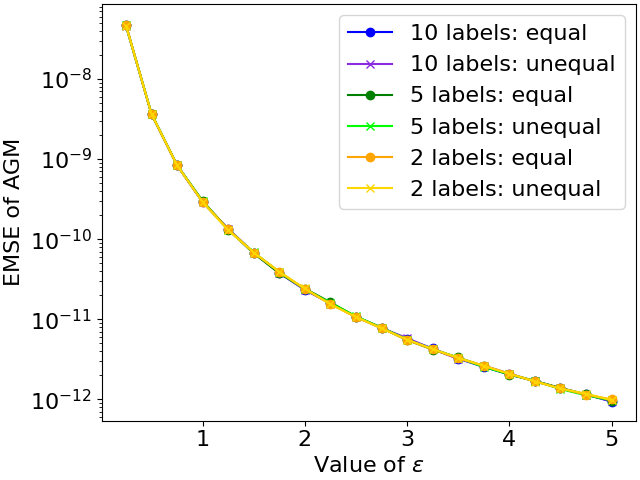}}
\quad
\subfloat[$Q$ for CIFAR-10]{\includegraphics[width=0.31\linewidth]{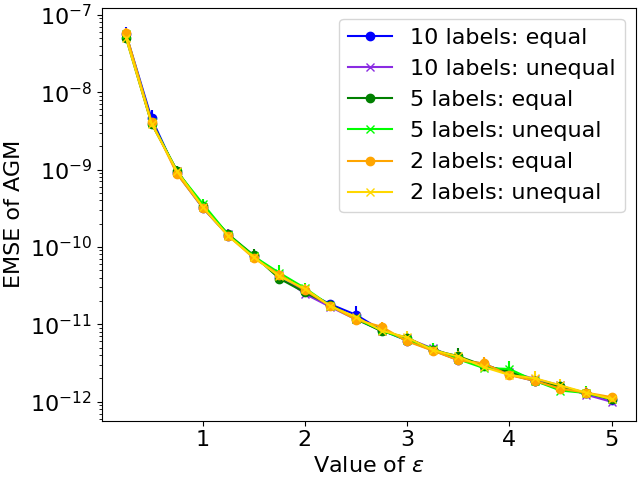}}
\quad
\subfloat[$I^{2}$ for CIFAR-10]{\includegraphics[width=0.31\linewidth]{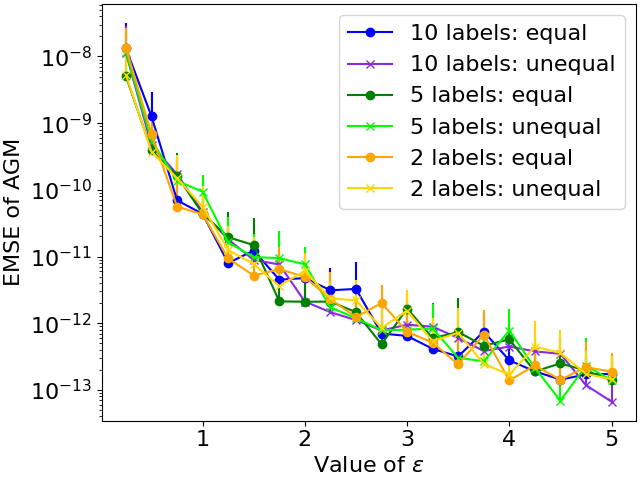}}
\quad
\subfloat[Dispersion for CIFAR-100]{\includegraphics[width=0.31\linewidth]{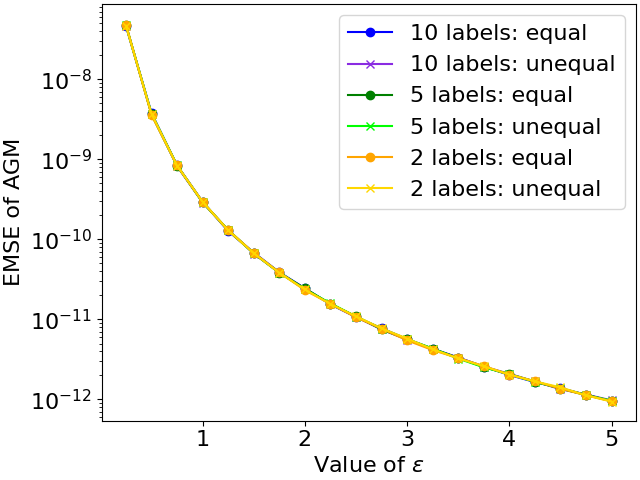}}
\quad
\subfloat[$Q$ for CIFAR-100]{\includegraphics[width=0.31\linewidth]{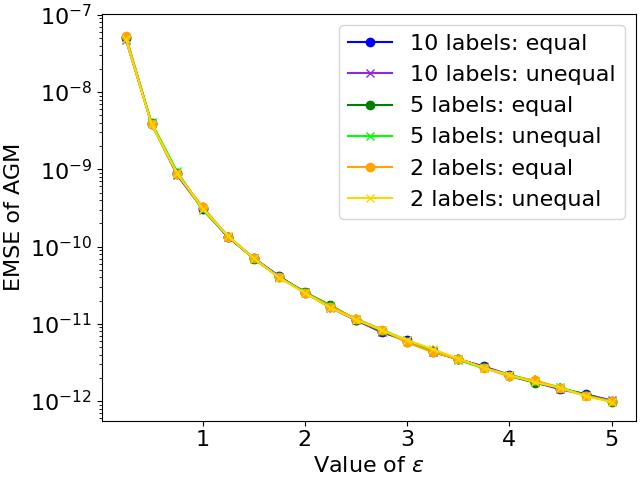}}
\quad
\subfloat[$I^{2}$ for CIFAR-100]{\includegraphics[width=0.31\linewidth]{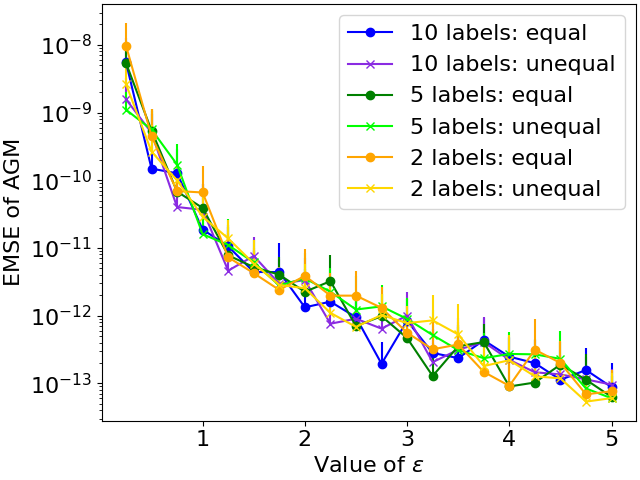}}
\quad
\subfloat[Disp. for Fashion-MNIST]{\includegraphics[width=0.31\linewidth]{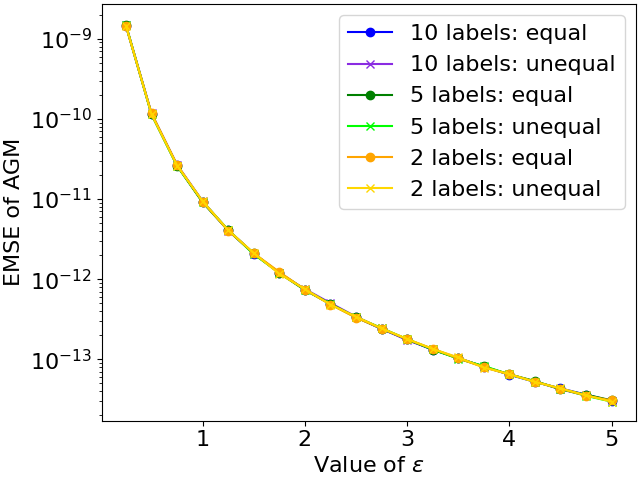}}
\quad
\subfloat[$Q$ for Fashion-MNIST]{\includegraphics[width=0.31\linewidth]{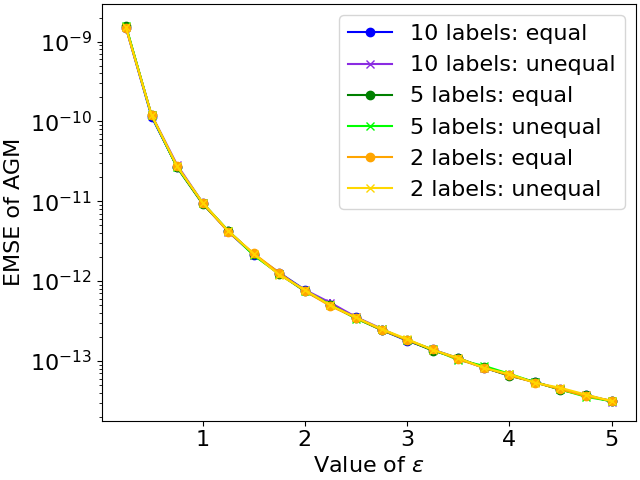}}
\quad
\subfloat[$I^{2}$ for Fashion-MNIST]{\includegraphics[width=0.31\linewidth]{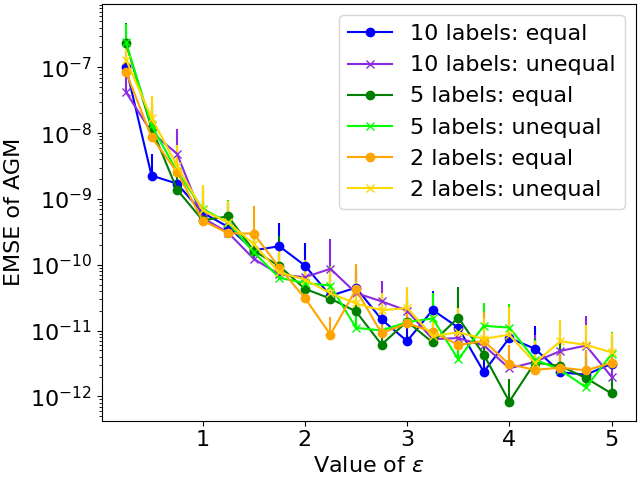}}
\caption{\small\label{fig:graphs} Effect of privacy parameter $\varepsilon$ on EMSE of AGM, depending on number of labels and statistical heterogeneity.}
\end{figure*}

Finally, we plot nine graphs in Figure~\ref{fig:graphs} to illustrate the relationship between the privacy parameter $\varepsilon$ and the EMSE, with each graph focusing on one measure of statistical heterogeneity combined with one dataset.
Each of the lines in each graph represents a particular level of statistical heterogeneity, as illustrated in the bullet points earlier in this subsection.
It can be clearly seen that as the value of $\varepsilon$ increases, the EMSE in all graphs decreases in a negative exponential fashion.
The negative trend occurs because a larger value of $\varepsilon$ implies that less noise is added, lowering the EMSE.
The exponential relationship is implied by Definition~\ref{defn:dp}.
Secondly, it can be seen that the variance of the levels of statistical heterogeneity is smallest when the dispersion metric is used, marginally increased when the $Q$ metric is used, and then much larger when the $I^{2}$ metric is used.
This is consistent with the literature because out of the three metrics only $I^{2}$ is able to evaluate the extent of statistical heterogeneity.
Finally, it can be observed that the EMSE is always smaller than $10^{-6}$, and therefore this level of noise does not significantly impact the percentage changes in Table~\ref{tab:changemse}.

\subsection{Evaluation}

Our experimental results have confirmed the main theoretical results in Section~\ref{sec:theory}, and have shown that the theoretical bounds are quite tight.
In Section~\ref{subsec:stats}, we demonstrated the superior accuracy of the AGM compared to the CGM, and evaluated the marginal advantage of our distributed setting compared with a centralized setting.

In Section~\ref{subsec:shprivacy}, having selected the optimal mechanism (AGM) and setting (distributed), we evaluate the percentage change in its accuracy, and the cost of achieving a particular level of privacy, at six different levels of heterogeneity.
We conclude that, out of the three measures of statistical heterogeneity, $I^{2}$ produces the best accuracy, but all measures do not lose significant accuracy when a heterogeneous sample is used.
In all cases, the noise levels do not encroach significantly on the accuracy.
Because the values of $I^{2}$ as reported by the AGM can be interpreted to be consistent with $I^{2}$, we can trust this metric to distinguish between the different levels of heterogeneity.

\section{Conclusions and future work} \label{sec:concfuture}

This document has presented a differentially private measure of the statistical heterogeneity for a dataset of $d$-dimensional vectors, collected in a distributed way from $n$ users, based on the $I^{2}$ measure from meta-analysis.
It is most efficient to introduce differential privacy to vectors via Gaussian noise, with dependency $\sqrt{d}$ instead of $d$ for an alternative.

Though the statistics $Q$ and $I^2$ were analyzed in the context of statistical heterogeneity, in future work it will be interesting to look at their main usage in meta-analysis.
In particular, what would be the impact of a meta-analysis approach on the theoretical and experimental outcomes of the same or a similar study?
In future research it may be possible to optimize the accuracy of the privacy parameters further, using R\'{e}nyi differential privacy~\cite{mironov}.
To go beyond the scope of this study, a comparison can be made between the above implementation and other non-i.i.d. settings, evaluating their simplicity, relevance and usefulness.

\appendix
\section{Proof of Corollary~\ref{cor:msedisp}}
\label{app:mse}
\primedispmse*

\begin{proof}
$D''(\vec{X}) - D'(\vec{X}) = \sum_{i=1}^{n} Y_{2}.$
\begin{align*}
D'(\vec{X}) - D(\vec{X}) &= \sum_{i=1}^{n} ((\vec{x}_{i} - \pvec{\mu}) - Y_{1})^{2} - \sum_{i=1}^{n} (\vec{x}_{i} - \pvec{\mu})^{2} \\
&= \sum_{i=1}^{n} \!\left[ (\vec{x}_{i} - \pvec{\mu})^{2} - 2Y_{1}(\vec{x}_{i} - \pvec{\mu}) + Y_{1}^{2} \right] - \sum_{i=1}^{n} \big[ (\vec{x}_{i} - \pvec{\mu}) \big]^{2} \\
&= \sum_{i=1}^{n} \!\left[ Y_{1}^{2} - 2Y_{1}(\vec{x}_{i} - \pvec{\mu}) \right] \\
&= \sum_{i=1}^{n} \left[ Y_{1} \big( Y_{1} - 2(\vec{x}_{i} - \pvec{\mu}) \big) \right].
\end{align*}
\end{proof}

\section{Proof of Lemma~\ref{lem:dispest}}
\label{app:est}
\primedispest*

\begin{proof} Using the properties $\sum_{i=1}^{n} \vec{x}_{i} = n\vec{\mu}$ and $\sum_{i=1}^{n} \pvec{\mu}^{2} = n\pvec{\mu}^{2}$, and Lemma~\ref{lem:noisymu}, $\textnormal{Var}\hspace{0.1em}(D''(\vec{X}))$ can be simplified to:
\begin{align*}
&\!\left[ \Big( \sum_{i=1}^{n} (\vec{x}_{i} - \pvec{\mu})^{2} \Big)/n - \Big( \sum_{i=1}^{n} (\vec{x}_{i} - \pvec{\mu}')^{2} \Big)/n \right]^{2} \\
&= \!\left[ \Big( \sum_{i=1}^{n} (\vec{x}_{i}^{\hspace{0.05em} 2} - 2\vec{\mu}\vec{x}_{i} + \pvec{\mu}^{2}) \Big)/n - \Big( \sum_{i=1}^{n} (\vec{x}_{i}^{\hspace{0.05em} 2} - 2\pvec{\mu}'\vec{x}_{i} + (\pvec{\mu}')^{2}) \Big)/n \right]^{2} \\
&= \!\left[ \Big( \sum_{i=1}^{n} \vec{x}_{i}^{\hspace{0.05em} 2} \Big)/n - 2n\pvec{\mu}^{2}/n + n\pvec{\mu}^{2}/n - \Big( \sum_{i=1}^{n} \vec{x}_{i}^{\hspace{0.05em} 2} \Big)/n + 2n\vec{\mu}\pvec{\mu}'/n - n(\pvec{\mu}')^{2}/n \right]^{2} \\
&= (-\pvec{\mu}^2 + 2\vec{\mu}\pvec{\mu}' - (\pvec{\mu}')^{2})^{2}
= (\vec{\mu} - \pvec{\mu}')^{4} = Y_{1}^{\hspace{0.05em} 4}.
\end{align*}
\end{proof}

\section{Proof of Theorems~\ref{thm:dispconf} and \ref{thm:isquaredconf}}
\label{app:conf}
\primedispconf*

\begin{proof}
Using the property that for the Gaussian random variable $Y_{1} \sim \mathcal{N}^{d}(\xi_{1}^{2})$, $\mathbb{E}[Y_{1}^{2n}] = (2n - 1)!!\hspace{0.1em}\xi_{1}^{\hspace{0.05em} 2n}$~\cite{papoulis}, the standard deviation $\sigma$ of the estimated average dispersion can be simplified to:
\begin{align*}
\sqrt{\mathbb{E}[(Y_{1}^{4})^{2}] - (\mathbb{E}[Y_{1}^{4}])^{2}}
&= \sqrt{\mathbb{E}[Y_{1}^{8}] - (3\hspace{0.1em}\xi_{1}^{\hspace{0.05em} 4})^{2}} \\
&= \sqrt{7!!\hspace{0.1em}\xi_{1}^{\hspace{0.05em} 8} - 9\hspace{0.1em}\xi_{1}^{\hspace{0.05em} 8}} = \sqrt{96\hspace{0.1em}\xi_{1}^{\hspace{0.05em} 8}} = 4\sqrt{6}\hspace{0.1em}\xi_{1}^{\hspace{0.05em} 4}.
\end{align*}

Inserting the simplification of $\textnormal{Var}\hspace{0.1em}(D''(\vec{X}))$ into the expression for the 95\% confidence interval along with its $z$-score of 1.96 gives the following expression:
\[
D(\vec{X}) = D''(\vec{X}) \pm 1.96 \cdot \frac{4\sqrt{6}\hspace{0.1em} \xi_{1}^{\hspace{0.05em} 4}}{\sqrt{n}} = D''(\vec{X}) \pm \frac{7.84\sqrt{6}(\xi_{1}^{\hspace{0.05em} 2})^{2}}{\sqrt{n}}.
\]
\end{proof}

\primeisquaredconf*

\begin{proof}
Using the expression $I^{2} = 1 - ((n - 1)/Q)$ and the property that for $Z_{1} \sim \mathcal{N}^{d}(\eta_{1}^{2})$, $\mathbb{E}[Z_{1}^{2n}] = (2n - 1)!!\hspace{0.1em}\eta_{1}^{\hspace{0.05em} 2n}$~\cite{papoulis}, the standard deviation $\sigma$ of the estimated average $I^{2}$ can be simplified to:
\begin{align*}
&\sqrt{\!\left( 1 - \frac{(n - 1)^{2}}{w_{i}^{2}\hspace{0.1em}\mathbb{E}[(Z_{1}^{4})^{2}]} \right)- \!\left( 1 - \frac{(n - 1)^{2}}{w_{i}^{2}(\mathbb{E}(Z_{1}^{\hspace{0.05em} 4}])^{2}} \right)} \\
&=
\sqrt{\frac{(n - 1)^{2}}{9w_{i}^{2} \eta_{1}^{8}} - \frac{(n - 1)^{2}}{7!!\hspace{0.1em}w_{i}^{2}\eta_{1}^{8}}} = \sqrt{\frac{32(n - 1)^{2}}{315w_{i}^{2}\eta_{1}^{8}}}.
\end{align*}

Inserting the simplification of $\textnormal{Var}\hspace{0.1em}((I^{2})'''(\vec{X}))$ into the expression for the 95\% confidence interval along with its $z$-score of 1.96 gives the following expression:
\begin{align*}
I^{2}(\vec{X}) &= (I^{2})'''(\vec{X}) \pm \sum_{i=1}^{n} \frac{1.96}{\sqrt{n}} \cdot \sqrt{\frac{32(n - 1)^{2}}{315\hspace{0.1em}w_{i}^{2}\eta_{1}^{8}}} \\ &= (I^{2})'''(\vec{X}) \pm \sum_{i=1}^{n} \frac{0.625(n - 1)}{w_{i}\sqrt{n}(\eta_{1}^{2})^{2}}.
\end{align*}
\end{proof}

\end{document}